\documentclass[prodmode,acmtist]{acmsmall} 

\usepackage{amsmath}
\usepackage{amssymb}

\makeatletter
\renewcommand{\maketag@@@}[1]{\hbox{\m@th\normalsize\normalfont#1}}%
\makeatother

\usepackage{algorithm}
\usepackage{algpseudocode}
\usepackage{amsmath}
\usepackage{graphics}
\usepackage{epsfig}

\usepackage{graphicx}

\usepackage{enumitem}
\usepackage{cite}
\usepackage{multirow}
\usepackage{makecell}
\usepackage{array}

\usepackage{inputenc}

\usepackage{dcolumn}

\usepackage{float}

\usepackage{color}

\usepackage{booktabs}
\usepackage{threeparttable}

\begin{document}

\markboth{C. Deng et al.}{Deep Multi-scale Discriminative Networks for Double JPEG Compression Forensics}

\title{Deep Multi-scale Discriminative Networks for Double JPEG Compression Forensics}
\author{CHENG DENG, ZHAO LI, and XINBO GAO
\affil{Xidian University}
DACHENG TAO
\affil{University of Sydney}}


\begin{abstract}
As JPEG is the most widely used image format, the importance of tampering detection for JPEG images in blind forensics is self-evident. In this area, extracting effective statistical characteristics from a JPEG image for classification remains a challenge. Effective features are designed manually in traditional methods, suggesting that extensive labor-consuming research and derivation is required. In this paper, we propose a novel image tampering detection method based on deep multi-scale discriminative networks (MSD-Nets). The multi-scale module is designed to automatically extract multiple features from the discrete cosine transform (DCT) coefficient histograms of the JPEG image. This module can capture the characteristic information in different scale spaces. In addition, a discriminative module is also utilized to improve the detection effect of the networks in those difficult situations when the first compression quality ($QF1$) is higher than the second one ($QF2$). A special network in this module is designed to distinguish the small statistical difference between authentic and tampered regions in these cases. Finally, a probability map can be obtained and the specific tampering area is located using the last classification results. Extensive experiments demonstrate the superiority of our proposed method in both quantitative and qualitative metrics when compared with state-of-the-art approaches.
\end{abstract}

%




\acmformat{Cheng Deng, Zhao Li, Dacheng Tao, and Xinbo Gao. 2018. Deep multi-scale discriminative networks for double JPEG compression forensics.}

\begin{bottomstuff}
This work was supported by the National Natural Science Foundation of China (Grant No. 61572388 and 61703327), and the Key R\&D Program-The Key Industry Innovation Chain of Shaanxi (Grant No. 2017ZDCXL-GY-05-04-02), and the Australian Research Council Projects (Grant No. FL-170100117, DP-180103424, and IH-180100002).

Author's addresses: C. Deng {and} Z. Li {and} X. Gao, School of Electronic Engineering, Xidian University, Xi'an 710071, China; emails: chdeng.xd@gmail.com, lizhao@stu.xidian.edu.cn, xbgao@mail.xidian.edu.cn; D. Tao, School of Information Technologies, Faculty of Engineering and Information Technologies, University of Sydney, J12/318 Cleveland St, Darlington NSW 2008, Australia; email: dacheng.tao@sydney.edu.au.
\end{bottomstuff}

\maketitle

\section{Introduction}

\label{sec:Introduction}

With the rapid development of image acquisition tools and the popularity of social media, digital images are now widely used and have become the major information carrier. Due to the variety of available image processing tools, people can easily modify an image in any way they want~\cite{li2017superpixel,li2017patch}. As a result, current digital technology has begun to erode trust in visual imagery in many fields, such as journalism, military, justice, commerce, medical applications, and academic research~\cite{a0,Liu2014Improved,Korus2016Multi,Chen2017Blind,dong2015texture}. Consequently, digital image forensics, which aims to identify the original source of an image or determine whether or not the content of an image has been modified, has become increasingly important.

Since JPEG is the image format used by most digital devices, research into JPEG-related forensics has attracted significant attention~\cite{Liu2011Neighboring,Liu2012,Thing2013}. JPEG compression identification on bitmaps and double compression detection on JPEG images are two main research topics in JPEG forensics. The goal of JPEG compression identification on bitmaps is to detect the tampering traces of an image that has been previously JPEG-compressed and stored in lossless format. Thai \emph{et al.}~\cite{thai2017jpeg} proposed an accurate method for estimating quantization steps from a lossless format image that has experienced JPEG compression. Yang \emph{et al.}~\cite{yang2015estimating} proposed a novel statistic named factor histogram for estimating the JPEG compression history of bitmaps. Li \emph{et al.}~\cite{li2015revealing} provided a novel quantization noise-based solution to reveal the traces of JPEG compression. In this paper, we focus on double compression forensics on JPEG images. Many forensics techniques are inapplicable to JPEG images because compression can weaken certain traces of image tampering. However, recompression often appears when the JPEG image is tampered with and re-saved in JPEG format~\cite{yang2014effective}. These processes will leave specific traces of double compression; consequently, many related methods aim to detect double JPEG compression from histograms of the Discrete Cosine Transform (DCT) coefficient~\cite{Nguyen2017}. Several works analyze or model the effect of JPEG compression. Yang \emph{et al.}~\cite{yang2016analyzing} presented a theoretical analysis on the variation of local variance caused by JPEG compression. Li \emph{et al.}~\cite{li2015statistical} presented a statistical analysis of JPEG noises, including the quantization noise and the rounding noise during a JPEG compression cycle.

\begin{figure}[!t]
\centering
\includegraphics[width=13.8 cm]{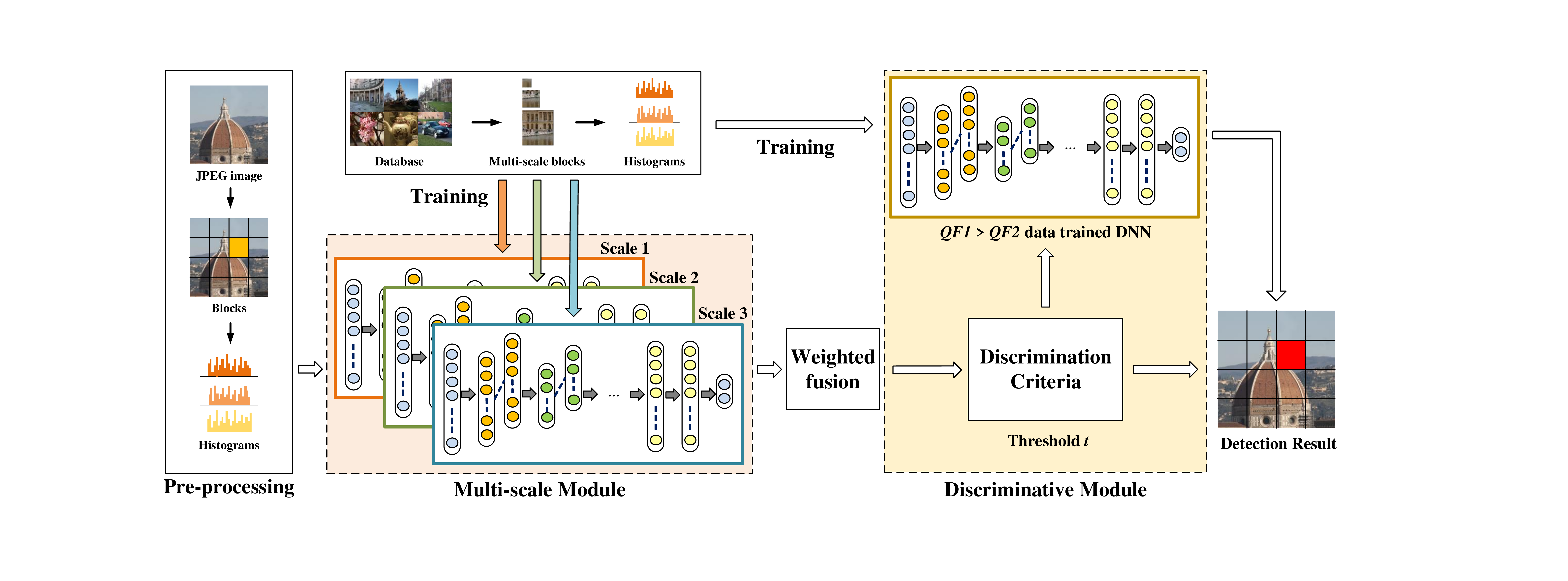}
\caption{Overview of our approach. The deep multi-scale discriminative networks contain a multi-scale module to extract features in different scale spaces and a discriminative module to judge whether another special network should be chosen for detection. Based on the final classification results, it is possible to determine whether the input image block is tampered or not.}
\label{fig:framework}
\end{figure}

The existing techniques for JPEG compression can be classified into two categories: the traditional method and the deep learning method. Many traditional algorithms for double JPEG compression have yielded relatively accurate detection results. Luk\'{a}\v{s} and Fridrich~\cite{a1} estimated the primary quantization matrix and presented a periodicity of DCT coefficient histograms due to double compression. Popescu and Farid~\cite{a2} put forward some statistical correlations caused by digital tampering and analyzed the Double Quantization (DQ) effect. However, these methods cannot locate specific area that has been tampered. Lin \emph{et al.}~\cite{a3} first proposed a fine-grained tampered image detection method that can locate the tampered region by investigating the DQ effect on DCT coefficients. Bianchi \emph{et al.}~\cite{a4} designed a probability map to distinguish between tampered and original regions. Bianchi and Piva~\cite{a5} proposed a kind of forensic algorithm to locate forged areas in JPEG format images by computing a likelihood map to represent the probability of each small DCT block being compressed once or twice. By modeling the DCT coefficient histograms as a mixture, the probability of blocks being tampered was obtained to locate tampered regions~\cite{a6}\cite{a7}. Liu~\cite{liu2017approach} utilized ensemble learning with nearly $100000$-dimensional features from the spatial domain and from the DCT transform domain to address the challenging detection problems when $QF1 \textgreater QF2$. Another type of traditional method is based on Benford's law. Fu \emph{et al.}~\cite{a8} utilized a Benford's law-based statistical model to distinguish between the tampered and authentic images, as the DCT coefficients of single compressed images obey the law but double compressed images do not. Li \emph{et al.}~\cite{a9} utilized mode-based first digit features to detect double compressed images and identify the primary quality factor of JPEG compression.

Almost all traditional methods require some artificially designed features for detection. However, designing such features is sometimes difficult and requires a large amount of theoretical research and experimentations. Another drawback is that traditional methods may disable detection when $QF1 \textgreater QF2$ due to the small difference between the statistical features of a tampered image and an authentic image in this case. Recently, deep neural network (DNN) in image processing and computer vision~\cite{Tang2015A,Yan2016Blind,Fu2016Clearing,Hu2017Visual} has been used with great success when applied to the image forensics field. Baroffio \emph{et al.}~\cite{a10} proposed a deep learning method to solve the problem of camera source identification. Chen \emph{et al.}~\cite{a11} utilized a convolutional neural network (CNN) to detect median filtering operations in images. Bayar and Stamm~\cite{a12} utilized a new convolutional layer to detect universal image manipulation and proposed a network structure based on actual forensic evidence. In double JPEG compression forensics, Wang and Zhang~\cite{a13} explored eight different CNNs to solve the tampering localization problem; in this study, eight networks were trained on single-compressed images and double-compressed images with different $QF2$. In an estimation of $QF2$, one corresponding network is selected to detect the tampered regions. However, this method extracts features directly without taking full account of some characteristics of the double JPEG compression process, limiting the improvement of the detection effect. Finally, this method does not propose a solution for tougher cases where $QF1 \textgreater QF2$, which is also a general difficulty for most existing methods. Amerini \emph{et al.}~\cite{Amerini2017Localization} explored the use of a spatial domain CNN and its combination with the histograms of DCT for the image forgery detection. The author suggested that further research is in progress. Barni \emph{et al.}~\cite{Barni2017Aligned} utilized CNNs in pixel domain, noise domain and DCT domain to perform detection task respectively. This method can obtain comprehensive information from a JPEG image but is quite complicated.

In this paper, we propose a novel method of deep multi-scale discriminative networks (MSD-Nets) for double JPEG compression forensics to detect tampered regions automatically. This generalization method does not require an estimation of $QF1$ and the networks can detect JPEG images with any $QF2$. The multi-scale module consists of a combination of three single-scale networks connected in parallel~\cite{Zhang2016}\cite{korus2017multi}. This module is designed to extract features from the histograms of DCT coefficient after preprocessing, and is better able to describe the effective features in different scale spaces than its single-scale counterparts. The outputs of these networks are fused with different weights. A discriminative module then follows behind to judge whether another specially designed network should be chosen for the detection task. This module can improve the ability of our networks to distinguish the small statistical difference between a tampered region and an authentic region when $QF1 \textgreater QF2$, which most existing methods can not do. After detecting all the blocks of a JPEG image, a specific probability map of the detection result is obtained and a precise localization of the tampered area is achieved. Fig.~\ref{fig:framework} visualizes the whole framework of our proposed method. The experimental results on the Synthetic and Florence datasets demonstrate that the proposed method outperforms state-of-the-art traditional methods and a representative deep method.

The rest of this paper is organized as follows. In Section II, the features to be extracted are introduced, after which the MSD-Nets for double JPEG compression forensics are presented in Section III. The experimental results, and analysis are included in Section IV. Finally, conclusions are drawn in Section V.

\section{Features in Double JPEG Compression}
\label{sec:Features}

In this section, the image tampering model and the DQ effect are first described. From these statistical characteristics, useful features to be employed in solving the binary classification problem of double JPEG compression forensics are extracted. Finally, the difference between DCT coefficient histograms in single and double compression situations is evaluated, showing the effectiveness of the features utilized in our networks.

\subsection{Image Tampering Model}
In the JPEG compression process, an input image is first divided into many $8 \times 8$ blocks, after which the discrete cosine transform is applied to every block. After the quantization process, a rounding function is applied to the DCT coefficients. These quantized DCT coefficients are later encoded via entropy encoding. The main reason for compression information loss lies in the quantization process~\cite{Thai2016}, whose quantization table is related to a particular compression quality factor from 0 to 100 in the form of an integer. As we know, a higher quality factor represents a lower amount of losses of image information, vice versa.

\begin{figure}[!t]
\centering
\includegraphics[width=8.8 cm]{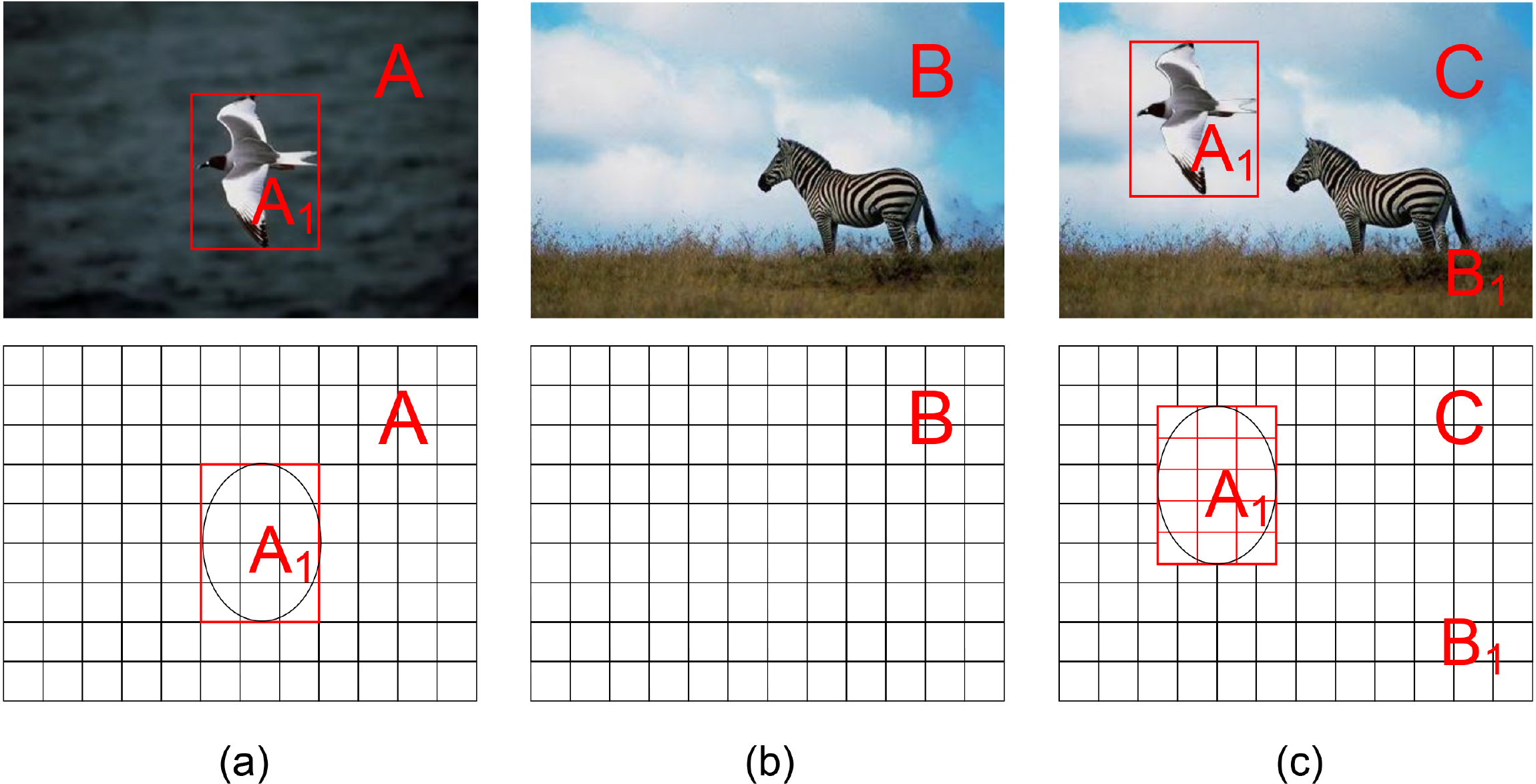}
\caption{Example of image tampering model: (a) source image $A$ with a selected region $A_1$, (b) original image $B$, (c) tampered image $C$.}
\label{fig:tamperingmodel}
\end{figure}

Image tampering is often accompanied by double JPEG compression. Traditionally, the process of image splicing includes three steps, as shown in Fig.~\ref{fig:tamperingmodel}. These steps are as follows:

1) Choosing a compressed JPEG image $B$, the quality factor of which is $QF1$, and decompressing it.

2) Replacing a region of $B$ with a selected region $A_1$ from another image $A$.

3) Saving the new tampered image $C$ in JPEG format with a quality factor $QF2$, where $B_1$ represents the authentic region of $B$.

In this model, $B_1$ is considered to be doubly compressed. If $A$ is in non-JPEG format, $A_1$ is undoubtedly singly compressed. However, if $A$ is in JPEG format, $A_1$ is regarded as a region that does not follow the law of double compression, as the DCT grids of $A_1$ have a very low probability $(\frac{1}{64})$ of matching $B_1$. Accordingly, $C$ is regarded as an image with a singly compressed region $A_1$ and a doubly compressed region $B_1$.

\subsection{Double Quantization Effect}
In the JPEG compression process, the main reason for information loss is quantization, which can leave traces from the histograms of the DCT coefficient. Here, the DQ effect causes periodic peaks and valleys in histograms after the process of double JPEG compression.

\begin{figure}[!t]
\centering
\includegraphics[width=13.8 cm]{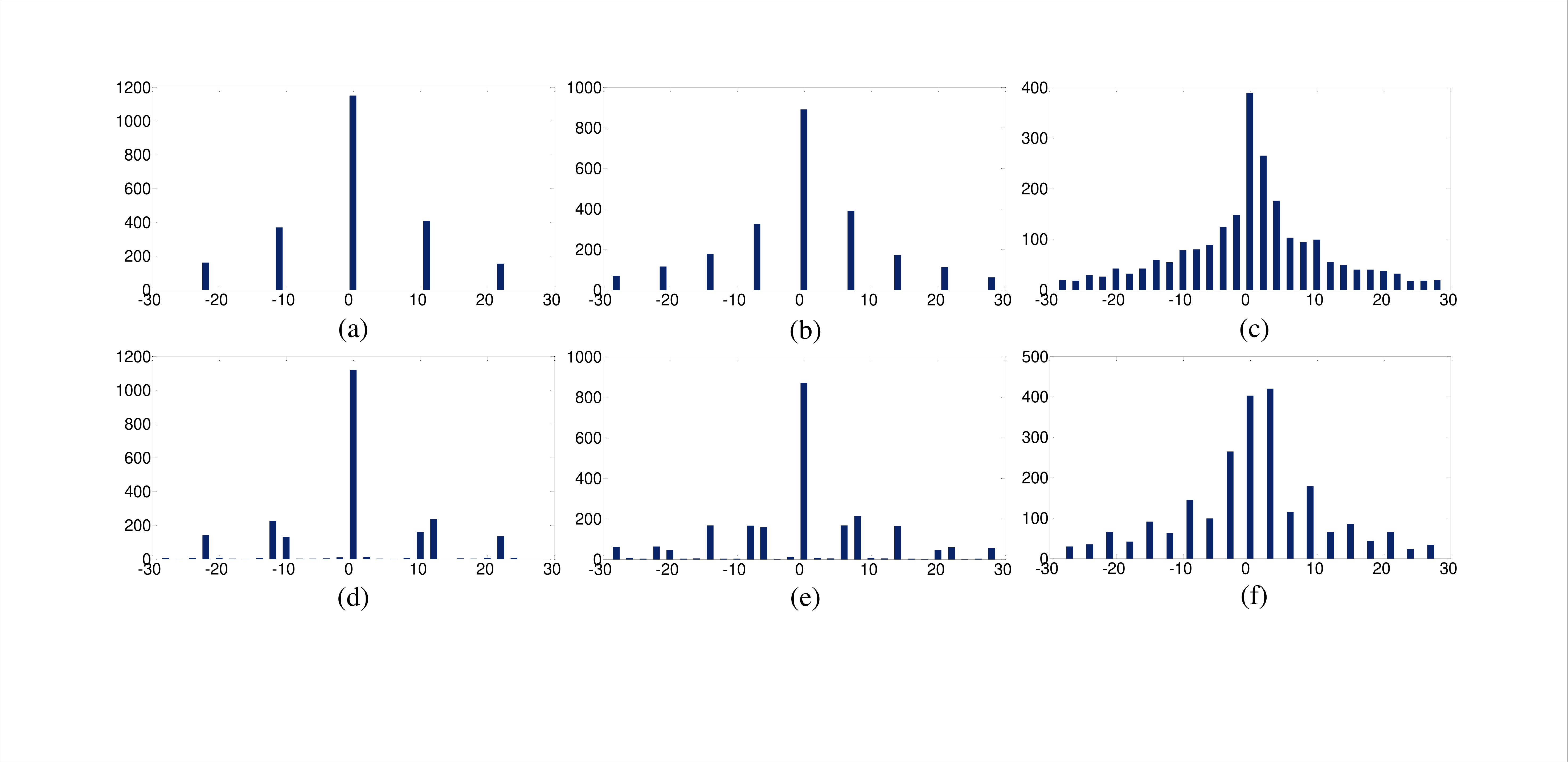}
\caption{Examples of different histograms in the DCT coefficients position (0,1): (a), (b), (c) single-compressed image when $QF1=50$, $QF1=70$, and $QF1=90$; (d), (e), (f) double-compressed image when (d) $QF1=50$, $QF2=90$, (e) $QF1=70$, $QF2=90$, and (f) $QF1=90$, $QF2=85$.}
\label{fig:histograms}
\end{figure}

Fig.~\ref{fig:histograms} illustrates the difference between double-compressed DCT coefficient histograms and single-compressed DCT coefficient histograms. From Fig.~\ref{fig:histograms}, we can see that the histograms of single-compressed images are generally in accordance with a generalized Gaussian distribution, but the histograms of double-compressed images have periodic peaks and valleys due to the DQ effect. When $QF1 \textgreater QF2$, the statistical properties of a double-compressed images are easily concealed by double compression. Thus, tampering detection becomes a tough problem in this situation.

To solve the problem of double JPEG compression forensics by classifying the singly and doubly compressed parts of a JPEG image, the DCT coefficient histograms are utilized as the input features.

\subsection{Tampering Detection Features}
In order to obtain the features that can be directly input into the networks, some pre-processing operations are needed. The DCT coefficients from the header file of a JPEG image are extracted first. The DCT coefficients contain a direct current (DC) coefficient and multiple alternating current (AC) coefficients. Taking an $8 \times 8$ DCT block as an example, the DC coefficient is the first number of the DCT coefficient matrix, and AC coefficients are the other 63 numbers. In this paper, only AC coefficients with distributions that are different from the DC coefficients are selected to remove the impact of DC coefficients.

\begin{figure}[!t]
\centering
\includegraphics[width=8.9 cm]{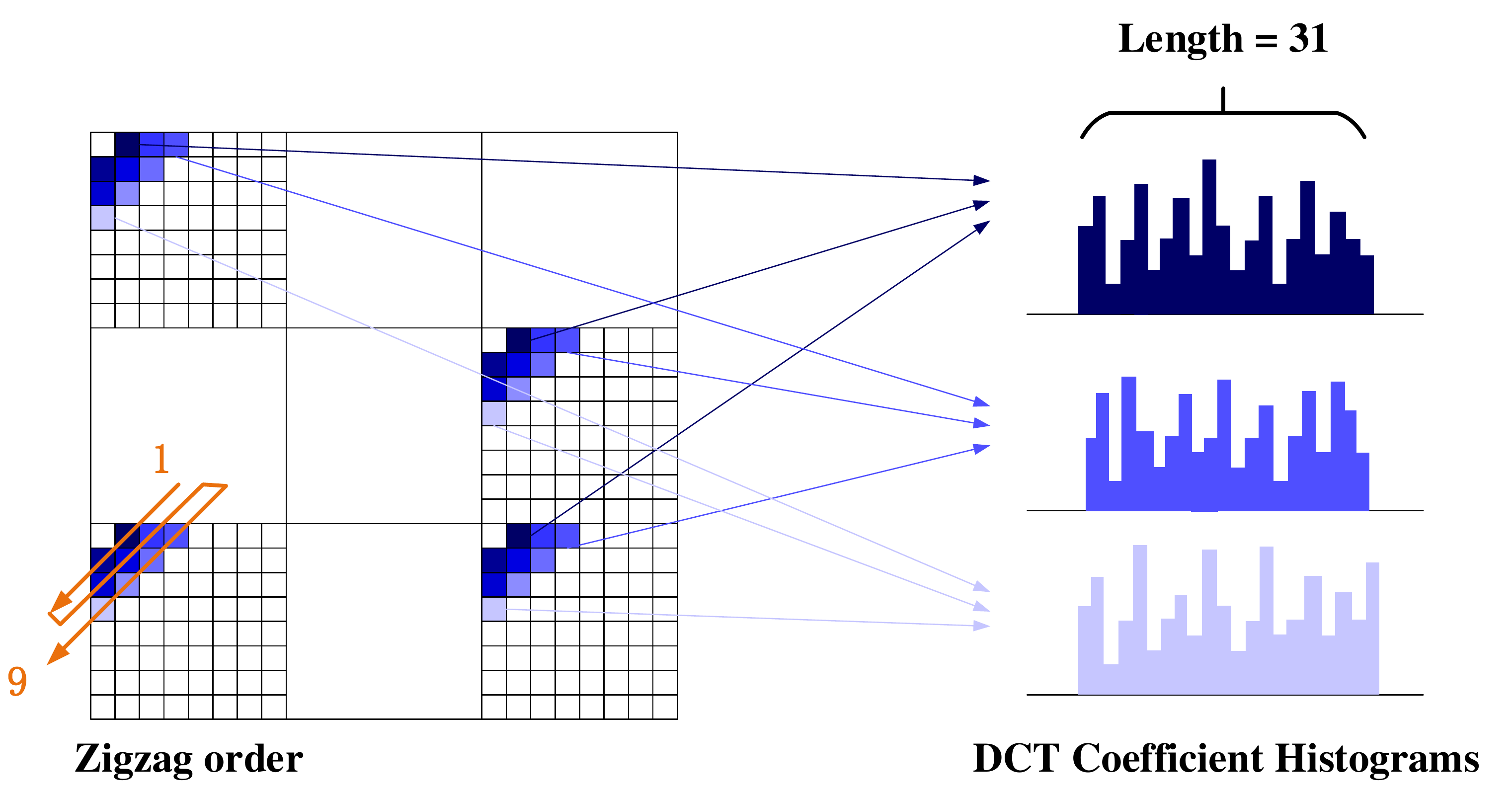}
\caption{The illustration of building histograms of DCT coefficients: (1) select the second to the tenth coefficients which are arranged in zigzag order, (2) extract the histograms with a length of 31 from each coefficient to construct the final features. Note only three blocks and three histograms are shown for succinctly explaining. Different colors are used to distinguish different blocks.}
\label{fig:DCTchangehistograms}
\end{figure}

Because of the variable sizes of the histograms, and for the purpose of controlling the computational consumption without losing significant information, a symmetric interval containing the peak of the histogram is selected as the features. Fig.~\ref{fig:DCTchangehistograms} shows a more detailed illustration. First, the second to the tenth coefficients, which are arranged in zigzag order, are chosen to organize the primary features. Then, the values of the positions $\{-15, -14, \ldots, 14, 15\}$ are utilized to construct the final features. If $F$ represents the feature set of a block from a JPEG image, and $H_i(x)$ represents the histogram of DCT coefficients corresponding to the position $x$ at the $i$th frequency form the zigzag order of the block, we have the following vector:

\begin{equation}
\begin{split}
F=\{ &H_i(-15), H_i(-14), \ldots, H_i(-2), H_i(-1),\\
&H_i(0),H_i(1),H_i(2),\ldots, H_i(14), H_i(15) \},\\
&i \in \{2,3,\ldots,9,10\}.
\end{split}
\end{equation}

Therefore, a 279 ($9 \times 31$) length feature vector is obtained from each JPEG image block. The parameter selection will be explained in the last part of Section IV.

\section{Tampering Detection via MSD-Nets}
In this section, our proposed model will be described in detail. During the preprocessing process, the histograms of DCT coefficient from a $64 \times 64$ block of a JPEG image are first extracted. After that, three networks trained by different scale data can automatically extract the different statistical features of the same histograms. Then, a discriminative module is employed to judge whether another special network should be chosen for detection, which is designed to improve classification accuracy for tougher cases when $QF1 \textgreater QF2$. A special network in the discriminative module can distinguish the small statistical difference between a tampered region and an authentic region in this case. Finally, a proposed localization scheme is utilized to obtain the probability map and simultaneously output the last two classification results.

\subsection{Network Architecture}
In order to solve the problem of double JPEG compression forensics, DNN is utilized to automatically extract the features from the DCT coefficient histograms and classify them. Our method does not need to estimate the first compression quality factor of the JPEG image, but instead utilizes a scheme of overall training. The image blocks, which have quality factors ranging from 50 to 95, are utilized for training to update the network parameters. In order to improve the effect of classification, four different DNNs are designed with similar structures but different parameters for training and testing.

The first part of MSD-Nets can be regarded as a three-channel structure for the process of extracting different features. This structure is inspired by the findings of a large number of experiments: namely, that the fusion of multi-scale networks trained by image blocks at different scales can increase the valuable information in various scale spaces for double JPEG compression tampering detection. This method of increasing the input features can introduce more effective information for classification, so that the characteristics of the network extraction are no longer limited to the single feature. This facilitates the extraction of more diversified and detailed information so that a better classification result can be obtained.

The pre-trained multiple DNNs can be utilized to extract the multi-scale features of DCT coefficient histograms automatically and features are aggregated through a process of weighted fusion. A three-scales model with three different networks, which are trained by $64 \times 64$, is designed using $128 \times 128$ and $256 \times 256$ size blocks respectively. Each kind of the block dataset consists of tampered and authentic blocks in the same quantity. The network trained by $64 \times 64$ blocks can extract features in small-scale space, while the network trained by $256 \times 256$ blocks can extract features in large-scale space, ensuring the richness of the feature in various scale spaces. The specific model selection will be explained in Section IV.

A variety of fusion methods have been tried, combining three kinds of features in the fully connected layer or using another special DNN to automatically select the weights. Finally, fixed weights have been found to yield better classification results. Therefore, the result-oriented fixed weights are utilized in our multi-scale feature DNN. The fusion process can be defined as:

\begin{equation}
\begin{split}
&S(1,1)=w_1*s_1(1,1)+w_2*s_2(1,1)+w_3*s_3(1,1),\\
&S(2,1)=w_1*s_1(2,1)+w_2*s_2(2,1)+w_3*s_3(2,1),\\
&s.t. \  w_1,w_2,w_3 \leqslant 1, w_1+w_2+w_3=1,
\end{split}
\end{equation}
where $S$ represents the result after the fusion process. $s_1$, $s_2$, and $s_3$ represent the output values after the softmax layer in three different networks. $w_1$, $w_2$, and $w_3$ represent the weights of $64 \times 64$, $128 \times 128$, and $256 \times 256$ blocks respectively.

\begin{table*}[!t] 
\tbl{The architecture of a single network structure.\label{tab:table2}}{ 
\renewcommand\arraystretch{1.5}
\centering
\tiny
\setlength{\tabcolsep}{1.3mm}{
\begin{tabular}{p{1.15cm}<{\centering}|p{0.55cm}<{\centering}|p{1cm}<{\centering}|p{1cm}<{\centering}|p{1cm}<{\centering}|p{0.9cm}<{\centering}|p{0.8cm}<{\centering}|p{0.65cm}<{\centering}|p{0.8cm}<{\centering}|p{0.65cm}<{\centering}|p{0.8cm}<{\centering}|p{0.6cm}<{\centering}|p{0.5cm}<{\centering}}
  \hline
  \hline
  Layer & Input & Conv1 & Pool1 & Conv2 & Pool2 & Full1 & ReLUs & Full2 & ReLUs & Full3 & Softmax & Output \\
  \hline
  Size & 279$\times$1 & 277$\times$1$\times$100 & 138$\times$1$\times$100 & 136$\times$1$\times$100 & 67$\times$1$\times$100 & 1000$\times$1 & 1000$\times$1 & 1000$\times$1 & 1000$\times$1 & 1000$\times$1 & 2$\times$1 & 2$\times$1 \\
  Kernel & - & 3$\times$1 & 3$\times$1 & 3$\times$1 & 3$\times$1 & - & - & - & - & - & - & - \\
  Stride & - & 1 & 2 & 1 & 2 & - & - & - & - & - & - & - \\
  Feature Map & - & 100 & - & 100 & - & 1000 & - & 1000 & - & - & - & - \\
  Property & - & - & Max & - & Max & - & - & - & - & - & - & - \\
  \hline
  Initialization
  (Weight, Bias) & - & Xavier Constant & - & Xavier Constant & - & Xavier Constant & - & Xavier Constant & - & Xavier Constant & - & - \\
  \hline
  \hline
\end{tabular}}}
\end{table*}

\begin{figure}[!t]
\centering
\includegraphics[width=13.8 cm]{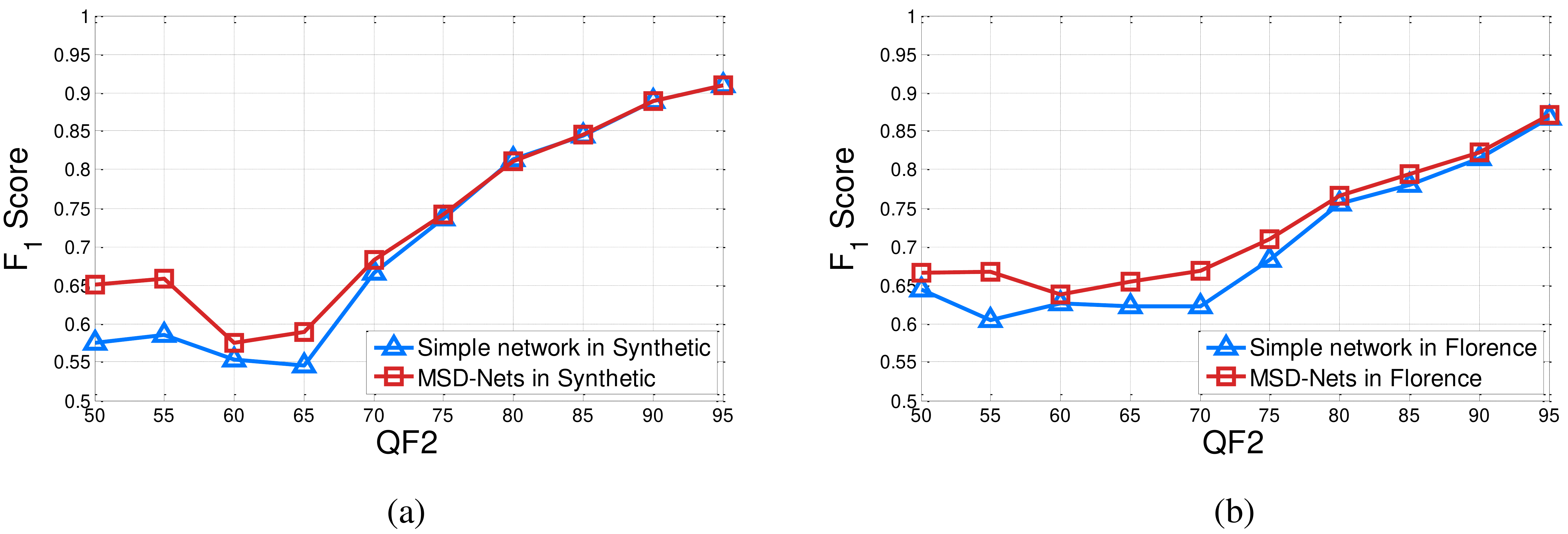}
\caption{$F_1\ Score$ of the proposed MSD-Nets structure and the simple network structure on (a) the Synthetic dataset, (b) the Florence dataset.}
\label{fig:multivssingle}
\end{figure}

In Fig.~\ref{fig:multivssingle}, $F_1$\emph{Score} is utilized to compare the experimental results of our MSD-Nets structure with a simple network on a 100000-block dataset. It is obvious that this specially designed structure for double JPEG compression forensics outperforms a simple network structure.

In addition, a divide and conquer strategy is implemented in our proposed method. Due to the statistical characteristics of the DQ effect, most of the existing schemes perform poorly when distinguishing the small statistical difference between a tampered region and an authentic region when $QF1\textgreater QF2$. Hence, some specific data ($64 \times 64$ tampered image blocks in the case of $QF1 \textgreater QF2$, and $64 \times 64$ image blocks without tampering) are utilized to train a special DNN in our algorithm framework. This training pattern ensures the effectiveness of the network when $QF1 \textgreater QF2$. The detailed discriminative module is described in Algorithm~\ref{alg::sequential}.

The architecture of a single channel network structure is shown in Table~\ref{tab:table2}. The network contains two alternating convolutional layers, two pooling layers and three fully connected layers. A softmax layer is utilized in the last of the structures to obtain the classification probability of each class.

\emph{1) Convolutional Layer:} A convolutional layer employs convolution and non-linearity operations to the input data, reduces the number of free parameters and simultaneously improves generalization~\cite{Lin2016}. A $3 \times 1$ size kernel is selected and the number of feature maps (that is, the number of kernels) is set to 100. The stride is set to 1. Hence, each feature map becomes a vector with a size of $277 \times 1$ and the output of the first convolutional layer becomes $277 \times 1 \times 100$. Similarly, the output of the second convolutional layer becomes $67 \times 1 \times 100$. The concrete convolutional operation is represented as:

\begin{equation}
x_j^l=\sum_{l=1}^n x_i^{l-1}*w_{ij}^{l-1}+b_j^l,
\end{equation}
where * represents convolution, $x_j^l$ is the $j$th feature map of layer $l$, $w_{ij}^{l-1}$ represents the trainable weight that connects the $i$th feature map of layer $l$$-1$ with the $j$th feature map of layer $l$, and $b_j^l$ represents the bias parameter of the $j$th feature map of layer $l$.

\emph{2) Pooling Layer:} While the extracted features can be utilized for classification after the convolutional layer, this may give rise to some challenges in calculation and be prone to over-fitting. Hence a pooling layer which can combine the outputs of neuron clusters into a single neuron is utilized~\cite{a16}~\cite{a17} and a max pooling selected to extract the maximum value from each of a cluster of neurons at the prior layer. The size of the pooling is $3 \times 1$ and the stride is 2. Thus, only the max value within the local area to the next layer is calculated.

\emph{3) Fully Connected Layer:} The fully connected layer connects each neuron in one layer to each neuron in another layer. The weights and the bias of the network can be adaptively renewed in the fully connected layers because of the error propagation procedure. Therefore, the last classification result will be fed back to automatically guide the feature extracting process, after which the learning mechanism can be set up \cite{a18}. In our network, the first two fully connected layers have 1000 outputs and the last one has 2 outputs.

\emph{4) ReLUs Nonlinearity:} ReLUs is the abbreviation of Rectified Linear Units. Following the first two fully connected layers, ReLUs is utilized because of its ability to facilitate fast convergence in some large models trained on some large datasets \cite{a19}. This layer applies the non-saturating activation function as:

\begin{equation}
f(x)=\max(x,0),
\end{equation}
where $x$ represents the input patch of the features.

Double JPEG compression tampering detection can be regarded as a two-classification problem: doubly compressed region (authentic region) and singly compressed region (tampered region). Hence, after a softmax layer, the classification probability of the two classes can be obtained. The parameter selection will be explained in Section IV.

Meanwhile, we compare our method with SVM classifier which is mentioned in~\cite{a9} by inputting the $279\times1$ histograms into SVM classifier. Then we find that SVM classifier has poor performance on these histograms. Most of the reason is that traditional machine learning techniques usually have no ability to process raw data. When the histograms are utilized for classification directly without handcrafted feature extraction, these techniques can hardly work. Hence, the actual benefits of deep network is to achieve representation learning automatically and capture important features easily for forensics.

\begin{algorithm}[!t]
\caption{Threshold-Based Discriminative Module Path Selection Algorithm}
\label{alg::sequential}
\begin{algorithmic}[1]
  \Require
    $S_0$: DCT coefficient histograms of input image blocks; $F(1,1)$: the first value of the result after fusion; $F(2,1)$: the second value of the result after fusion; $C$: output value of the special DNN; $t$: Threshold;
\Ensure
Result of whether the block is tampered or not;
\If{$\ \left|{F(1,1)-F(2,1)}\right|\textless t\ $}
\State Compute the output $C$ of the special DNN with the input of $S_0$; Set $C$ as the final result;
\If{$\ C(1,1) \textgreater C(2,1)\ $}
\State The input block was manipulated.
\Else
\State The input block is authentic.
\EndIf
\Else
\State Set $F$ as the final result;
\If{$\ F(1,1) \textgreater F(2,1)\ $}
\State The input block was manipulated.
\Else
\State The input block is authentic.
\EndIf
\EndIf
\end{algorithmic}
\end{algorithm}

\subsection{Tampered Region Localization}
In order to locate the tampered area more precisely, a $M \times N$ input JPEG image $I$ is first divided into many $L \times L$ overlapping blocks ($L$ is set to 64 according to testing). We then compute the DCT coefficient histograms $H$ with a size of $279\times1$ from each block, and input $H$ to the MSD-Nets. The final probability pair is $P(1,1)$ and $P(2,1)$ after the softmax layer is obtained later. Finally, the value $P(1,1)$, which represents the probability that the block is singly compressed (i.e. tampered), will be assigned to the small $8 \times 8$ block in its center. Each overlapping block from image $I$ has a small $8 \times 8$ block in its center, so the number $n$ of these overlapping blocks from image $I$ is the same as the number of these small central blocks $n_c$. Therefore, $n$ can be computed as:

\begin{equation}
n=n_c=\left(\left\lfloor\frac{M-L}{8}\right\rfloor+1\right)*\left(\left\lfloor\frac{N-L}{8}\right\rfloor+1\right).
\end{equation}
~\par

After assigning all of the $L \times L$ blocks, a tampering detection probability map $P_{map}$ with a size of $8 \times n$ is obtained. Finally, the pixel values of the blocks on the edge of $P_{map}$ are padded as 0 and a $M \times N$ tampering detection result map is obtained. The whiter areas represent the higher probability that this particular block is tampered. It is natural that the binary classification result map can be easily obtained with a threshold of $0.5$.

\section{Experimental Results}
In this section, the results of extensive experiments in both quantitative and qualitative metrics are provided and compared. We compare our method with two representative traditional methods~\cite{a5}~\cite{a7} and a deep method \cite{a13}, denoted respectively as BP, YH, and WZ in plot legends. In addition, we compare our method with a traditional method, named LA~\cite{liu2017approach}. For fair comparison, we run BP, YH, and WZ according to their public codes, and reimplement the algorithm of LA based on the original paper.

\subsection{Dataset}
\textbf{UCID Dataset}. One of the most widespread lossless datasets in image forensics: the Uncompressed Color Image Database (UCID)~\cite{a14}. The UCID dataset contains 1338 images in TIFF format with a resolution of $512 \times 384$.

\textbf{Synthetic Dataset}. This dataset was synthesized from UCID dataset. 1200 images from UCID dataset are randomly selected for experiments (800 as training sets, 200 as validating sets and 200 as testing sets). To create JPEG images for training, validating, and testing, each original TIFF image is first compressed with a given quality factor $QF1\in\{50,55,\dots,90,95\}$. The left $1/2$ region that should not be doubly compressed is then replaced with the corresponding region in the original image. Finally, the image is compressed again with another quality factor $QF2$$\in\{50,55,\dots,90,95\}$. Each image achieves 100 possible combinations of $(QF1,QF2)$.

\renewcommand{\arraystretch}{1.25} 
\begin{table*}[!t]
  \centering
  \fontsize{8}{7.3}\selectfont
  \begin{threeparttable}
  \tbl{\textcolor[rgb]{0,0,0}{Detection accuracy achieved on the Synthetic dataset.}\label{tab:Synthetic_comparison}}{ 
    \setlength{\tabcolsep}{1.3mm}{
    \begin{tabular}{cccccccccccc}
    \toprule
    \multirow{2}{*}{$QF1$}& \multirow{2}{*}{Method}&
    \multicolumn{9}{c}{$QF2$}\cr
    \cmidrule(lr){3-12} & &50&55&60&65&70&75&80&85&90&95\cr
    \midrule
    \multirow{4}{*}{50}&Proposed&\bf0.5391&\bf0.5251&0.5476&\bf0.7378&\bf0.9194&\bf0.9557&\bf0.9804&\bf0.9808&\bf0.9767&0.9758\cr
    ~&BP&0.4756&0.5023&\bf0.6136&0.7021&0.8955&0.8512&0.8981&0.9375&0.9339&0.9222\cr
    ~&YH&0.5085&0.5111&0.5970&0.7080&0.8704&0.9225&0.8926&0.9020&0.9349&0.9170\cr
    ~&WZ&0.5167&0.5080&0.5082&0.5048&0.4980&0.6126&0.7793&0.8466&0.9122&0.9365\cr
    ~&LA&0.4986&0.5230&0.5653&0.7108&0.8223&0.8684&0.9217&0.9308&0.9455&\bf0.9759\cr
    \hline
    \multirow{4}{*}{60}&Proposed&0.5118&0.5040&\bf0.5769&0.5150&\bf0.7348&\bf0.9109&\bf0.9806&\bf0.9769&\bf0.9651&\bf0.9880\cr
    ~&BP&0.5195&0.4782&0.4300&0.5332&0.6748&0.8330&0.8571&0.8949&0.9264&0.9206\cr
    ~&YH&0.5160&0.5026&0.4967&0.5531&0.7129&0.7259&0.9131&0.8910&0.9382&0.9144\cr
    ~&WZ&0.5094&0.5061&0.5054&0.5028&0.5065&0.6689&0.8104&0.8238&0.9138&0.9332\cr
    ~&LA&\bf0.5323&\bf0.5375&0.5000&\bf0.5628&0.6463&0.7557&0.8233&0.8724&0.9205&0.9603\cr
    \hline
    \multirow{4}{*}{70}&Proposed&0.4875&0.5328&\bf0.6194&0.5204&\bf0.5097&\bf0.5902&\bf0.9147&\bf0.9535&\bf0.9587&\bf0.9746\cr
    ~&BP&0.5804&0.4980&0.4736&0.4893&0.4473&0.5029&0.6787&0.8125&0.8893&0.9163\cr
    ~&YH&\bf0.5947&0.5332&0.4844&0.5000&0.4788&0.5137&0.7705&0.9098&0.9320&0.9242\cr
    ~&WZ&0.5171&0.5091&0.5189&0.5054&0.5024&0.5083&0.7341&0.8542&0.8905&0.9284\cr
    ~&LA&0.5810&\bf0.6053&0.5799&\bf0.5433&0.5010&0.5822&0.7410&0.8508&0.8995&0.9493\cr
    \hline
    \multirow{4}{*}{80}&Proposed&0.5494&0.4947&0.6224&\bf0.6463&\bf0.7456&\bf0.6822&\bf0.5362&\bf0.7614&\bf0.9565&0.8941\cr
    ~&BP&0.5000&0.5000&0.4993&0.5407&0.5000&0.4945&0.4622&0.5358&0.7331&0.8887\cr
    ~&YH&0.5029&0.4993&0.5000&0.5101&0.5020&0.5007&0.4919&0.5306&0.8158&\bf0.9261\cr
    ~&WZ&0.5011&0.5042&0.5061&0.5031&0.5017&0.5258&0.5232&0.6737&0.8535&0.8548\cr
    ~&LA&\bf0.6162&\bf0.6151&\bf0.6509&0.6415&0.6353&0.6021&0.4994&0.6810&0.8220&0.8958\cr
    \hline
    \multirow{4}{*}{90}&Proposed&0.4862&0.5754&0.5277&\bf0.5647&\bf0.5527&0.5150&0.5992&\bf0.6936&0.5207&0.7344\cr
    ~&BP&0.4694&0.5156&0.4290&0.4632&0.4951&0.5400&0.5020&0.5000&0.4987&\bf0.7829\cr
    ~&YH&0.5000&0.5000&0.5026&0.4935&0.5055&0.5205&0.4941&0.5000&0.5000&0.7738\cr
    ~&WZ&0.5088&0.4990&0.5013&0.5061&0.5104&0.4955&0.5526&0.5931&0.5153&0.7662\cr
    ~&LA&\bf0.5936&\bf0.5768&\bf0.5895&0.5470&0.5419&\bf0.6094&\bf0.6711&0.5566&\bf0.7012&0.7792\cr
    \bottomrule
    \end{tabular}}}
    \end{threeparttable}
\end{table*}

\renewcommand{\arraystretch}{1.25} 
\begin{table}[!t]
  \centering
  \fontsize{8}{7.3}\selectfont
  \begin{threeparttable}
  \tbl{\textcolor[rgb]{0,0,0}{Detection accuracy achieved on the Florence dataset.}\label{tab:Florence_comparison}}{ 
    \setlength{\tabcolsep}{1.3mm}{
    \begin{tabular}{cccccccccccc}
    \toprule
    \multirow{2}{*}{$QF1$}& \multirow{2}{*}{Method}&
    \multicolumn{9}{c}{$QF2$}\cr
    \cmidrule(lr){3-12} & &50&55&60&65&70&75&80&85&90&95\cr
    \midrule
    \multirow{4}{*}{50}&Proposed&\bf0.5328&    0.5553&    0.5701&    \bf0.7775&    0.8635&    0.8996&    0.9174&    0.9158&    0.9298&    0.9519\cr
    ~&BP&0.5023&    \bf0.5739&    \bf0.6587&    0.7004&    0.8532&    \bf0.9071&    \bf0.9318&    0.9407&    0.9553&    0.9148\cr
    ~&YH&0.5204&    0.3984&    0.5215&    0.6366&    0.8373&    0.9003&    0.9310&    \bf0.9416&    \bf0.9563&    0.9200\cr
    ~&WZ&0.5178&    0.4997&    0.4728&    0.4507&    0.4288&    0.6422&    0.7071&    0.6587&    0.7222&    \bf0.9618\cr
    ~&LA&0.5204&    0.5517&    0.6423&    0.7648&    \bf0.8654&    0.8815&    0.8977&    0.9090&    0.9129&    0.9225\cr

    \hline
    \multirow{4}{*}{60}&Proposed&0.5502&    0.5705&    \bf0.5791&    \bf0.6008&    \bf0.7981&    \bf0.9010&    \bf0.9106&    \bf0.9129&    \bf0.9361&    0.9463\cr
    ~&BP&0.5747&    0.5401&    0.4992&    0.5975&    0.7027&    0.7661&    0.8933&    0.9111&    0.9175&    0.8733\cr
    ~&YH&\bf0.6018&    \bf0.6021&    0.5137&    0.5355&    0.6295&    0.7345&    0.8871&    0.9111&    0.9218&    0.8851\cr
    ~&WZ&0.5445&    0.5338&    0.5135&    0.4901&    0.4599&    0.6936&    0.7514&    0.6938&    0.7909&    \bf0.9819\cr
    ~&LA&0.5446&    0.5240&    0.5104&    0.5794&    0.6910&    0.8271&    0.8448&    0.8771&    0.8865&    0.8975\cr
    \hline
    \multirow{4}{*}{70}&Proposed&0.5586&    0.5981&    \bf0.7525&    \bf0.6176&    \bf0.5863&    \bf0.7130&    0.9113&    \bf0.9239&    \bf0.9229&    0.9526\cr
    ~&BP&0.5911&    0.6063&    0.5938&    0.5336&    0.5031&    0.5912&    0.8052&    0.8815&    0.9156&    0.8691\cr
    ~&YH&0.5996&    0.6080&    0.6262&    0.5883&    0.4843&    0.5841&    0.7771&    0.8744&    0.9175&    0.8812\cr
    ~&WZ&0.5969&    0.5860&    0.5631&    0.5404&    0.5186&    0.4966&    \bf0.9442&    0.7601&    0.8220&    \bf0.9824\cr
    ~&LA&\bf0.6802&    \bf0.6492&    0.5818&    0.5292&    0.5179&    0.5817&    0.7608&    0.8198&    0.8587&    0.8756\cr
    \hline
    \multirow{4}{*}{80}&Proposed&0.5375&    0.5272&    \bf0.7905&    \bf0.7521&    \bf0.8448&    \bf0.8236&    \bf0.5569&    \bf0.8996&    \bf0.9253&    0.9456\cr
    ~&BP&0.5265&    0.5467&    0.5278&    0.5505&    0.5782&    0.5457&    0.4996&    0.6449&    0.8871&    0.8572\cr
    ~&YH&0.5075&    0.5716&    0.5508&    0.5488&    0.5936&    0.5856&    0.4878&    0.5809&    0.8814&    0.8690\cr
    ~&WZ&0.4816&    0.4720&    0.6495&    0.5841&    0.5635&    0.5576&    0.5251&   0.8560&    0.9240&    \bf0.9666\cr
    ~&LA&\bf0.5813&    \bf0.5881&    0.7302&    0.7237&    0.6652&    0.5767&    0.5229&    0.6777&    0.7952&    0.8288\cr
    \hline
    \multirow{4}{*}{90}&Proposed&\bf0.5507&    \bf0.5531&    \bf0.6910&    \bf0.6634&    \bf0.6937&    0.6183&    \bf0.7692&    \bf0.8587&    0.5482&    0.9169\cr
    ~&BP&0.5083&    0.5053&    0.5057&    0.5034&    0.5169&    0.5112&    0.5297&    0.5557&    0.5002&    0.6473\cr
    ~&YH&0.5113&    0.5138&    0.4926&    0.4455&    0.5466&    \bf0.6271&    0.5104&    0.6017&    \bf0.5708&    0.7509\cr
    ~&WZ&0.5029&    0.5190&    0.5561&    0.5209&    0.4959&   0.5255&    0.6884&    0.7406&    0.5021&    \bf0.9721\cr
    ~&LA&0.5188&    0.5375&    0.5398&    0.5162&    0.5475&    0.5594&    0.6973&    0.6606&    0.5142&    0.7021\cr
    \bottomrule
    \end{tabular}}}
    \end{threeparttable}
\end{table}

\textbf{Florence Dataset}\footnote{ftp://lesc.dinfo.unifi.it/pub/Public/JPEGloc.}. The Image Dataset for Localization of Double JPEG compression (Florence Dataset) is a public dataset containing 100 full-resolution raw color images from three different digital cameras: Nikon D90, Canon EOS 450D, and Canon EOS 5D. These are converted to TIFF format images and compressed by $1/2$, $1/16$, and $15/16$ in this dataset. Only a $1024 \times 1024$ region in the central position of each image is utilized. The $1/2$ compressed dataset is chosen for experiments (64 for training, 16 for validating and 20 for testing). This dataset is also utilized in the experiments of~\cite{a5}.

Finally, $1000 \times 100$ low-resolution synthesized JPEG images from the Synthetic dataset and $100 \times 100$ high-resolution synthesized JPEG images from the Florence dataset are obtained. The left half of each JPEG image is singly compressed, providing a convenient comparison process for experiments. This measure of processing data facilitates to gain balanced samples from the same image in the following steps.

\subsection{Quantitative Experiments}
After generating a set of specific compressed images using the Synthetic dataset, cropping is first performed to divide each image into many $64 \times 64$ blocks, and 48 blocks on the Synthetic dataset are obtained. Hence, a positive set with $24 \times 1000 \times 100$ elements and a negative set with the same number of elements on the Synthetic dataset are obtained. Similarly, the number changes to $128 \times 80 \times 100$ on the Florence dataset, and $128 \times 128$ and $256 \times 256$ block datasets are obtained in the similar way. The dataset utilized in our discriminative module is rather special, as it consists of two parts: $64 \times 64$ singly compressed blocks with $QF1 > QF2$, and $64 \times 64$ doubly compressed blocks.

$80\%$ of the data is utilized for network training, with the remaining $20\%$ being used for validating. Three multi-scale feature DNNs and one special feature DNN are trained. The very popular Caffe implementation~\cite{a20} is utilized for the training task. Because of the huge computation complexity of the network, our experiments utilize NVIDIA GTX TITAN X to accelerate the process. The optimization method we used is Stochastic Gradient Descent. The value of the learning rate is 0.0005. The batch-size is 200, and the momentum is set to 0.9. The number of epochs is set to 20 to ensure network convergence.

For testing, the rest of the 200 images in the Synthetic dataset and 20 images in the Florence dataset are utilized. After dividing each image into overlapping $64 \times 64$ blocks with a stride of 8, these blocks are input into the MSD-Nets. In the weighted fusion step, the weight $w_1$ of $64 \times 64$ blocks is set to $0.8$, the weight $w_2$ of $128 \times 128$ blocks is set to $0.1$, and the weight $w_3$ of $256 \times 256$ blocks is set to $w_3=1-w_1-w_2=0.1$. Subsequently, the final result map is obtained.

\begin{figure}[!t]
\centering
\includegraphics[width=0.90\textwidth]{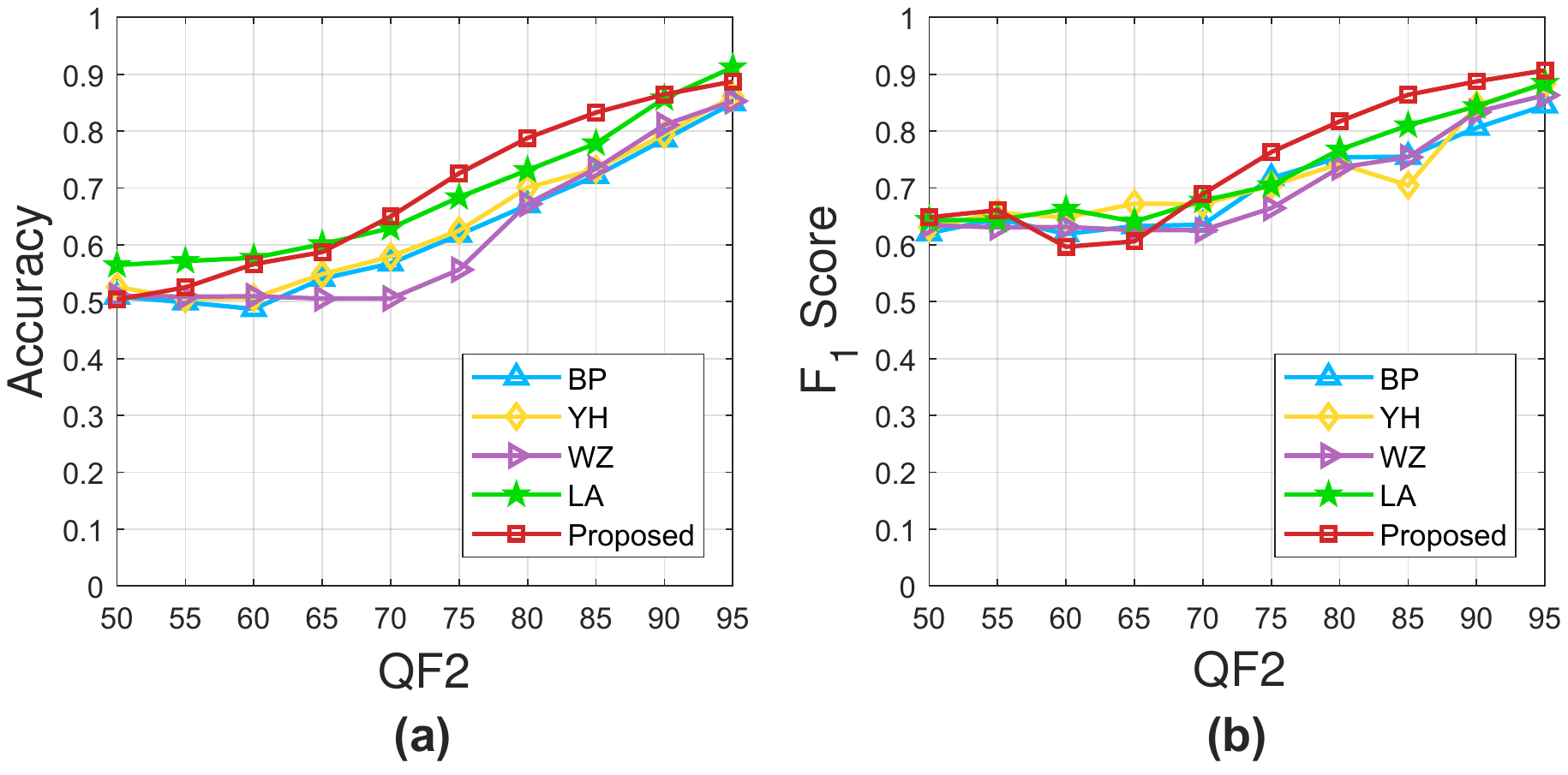}
\caption{\textcolor[rgb]{0,0,0}{Results achieved on the Synthetic dataset by the proposed method and the comparison methods of BP, YH, WZ, and LA. (a)-(b) average values of the detection accuracy and $F_{1}Score$ corresponding to $QF2$.}}
\label{fig:accf1s_UCID}
\end{figure}

\begin{figure}[!t]
\centering
\includegraphics[width=0.90\textwidth]{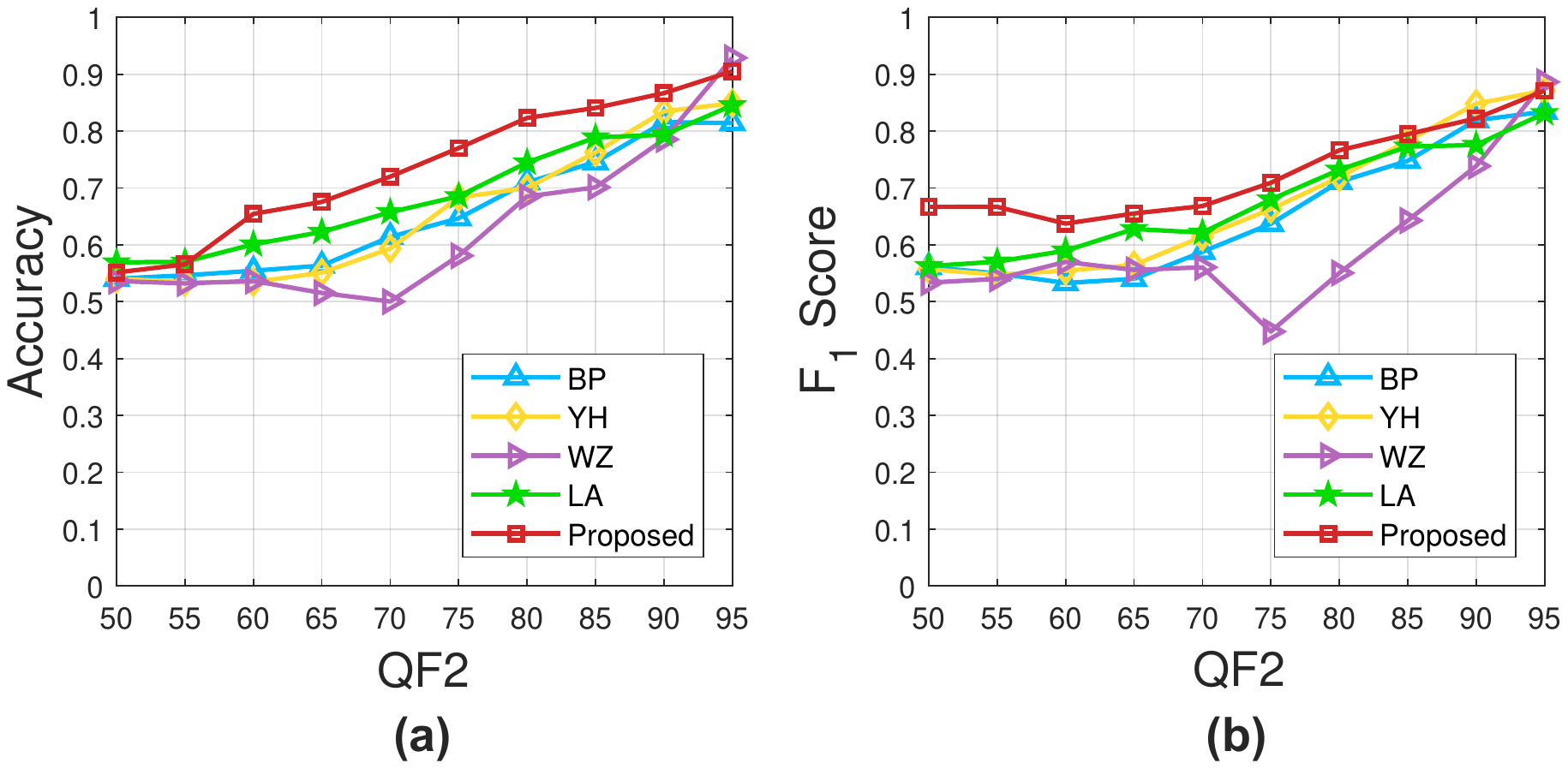}
\caption{\textcolor[rgb]{0,0,0}{Results achieved on the Florence dataset by the proposed method and the comparison methods of BP, YH, WZ, and LA. (a)-(b) average values of the detection accuracy and $F_{1}Score$ corresponding to $QF2$.}}
\label{fig:accf1s_Dresden}
\end{figure}

For quantitative experiments, the output probability map of the MSD-Nets is binarized to generate the final result map with two regions: a single-compressed region and a double-compressed region. Therefore, a pixel with a value of 0 represents being doubly compressed, while a value of 1 represents being singly compressed.

Accordingly, the metrics accuracy ($Acc$) and $F_{1}Score$ can be measured as:

\begin{equation}
Acc=\frac{TP+TN}{P+N},
\end{equation}

\begin{equation}
precision=\frac{TP}{TP+FP},
\end{equation}

\begin{equation}
recall=\frac{TP}{TP+FN},
\end{equation}

\begin{equation}
F_{1}Score=\frac{2 \cdot precision \cdot recall}{precision + recall },
\end{equation}
where $N$ is the total number of the single-compressed blocks, and $P$ is the total number of the double-compressed blocks. True positive $TP$ and true negative $TN$ stand for the numbers of blocks in double-compressed and single-compressed regions that are correctly identified. $FP$ and $FN$ denote the number of blocks which are erroneously detected as double-compressed blocks and the number of blocks that are falsely detected as single-compressed blocks. $F_{1}Score$ is a comprehensive indicator of $precision$ and $recall$.

The detection accuracy achieved on the Synthetic dataset and the Florence dataset is shown in Table~\ref{tab:Synthetic_comparison} and Table~\ref{tab:Florence_comparison}. In order to ensure the integrity of the experiment, the situation of $QF1=QF2$ is retained.

Table~\ref{tab:Synthetic_comparison} shows that the detection accuracies of our method are mostly $5$ to $20$ percent higher than BP, YH and WZ in the case of $QF1 \geq QF2$, especially when $QF1\in\{80,90\}$. When $QF1 \geq QF2$, the performance of our method are almost comparable to LA that is specially designed for down-recompression discrimination. In almost all cases when $QF1 < QF2$, the detection accuracies of our method are $5$ to $25$ percent higher than other methods. Similarly, according to Table~\ref{tab:Florence_comparison}, the detection accuracies of our method are $5$ to $15$ percent higher than other methods, even higher than LA, in most cases of $QF1 \geq QF2$. When $QF2\in\{60,\dots,90\}$, the superiority of our method is more prominent. In most cases of $QF1 < QF2$, the detection accuracies of our method are $5$ to $25$ percent higher than other methods, and the superior performances of our method are especially evident when $QF1\in\{60,80\}$. Generally, depending on the superior capability of the multi-scale module in feature extraction and the application of the discriminative module to address the challenging detection problems when $QF1 > QF2$, our method gains significantly higher accuracy when $QF1 > QF2$ as well as $QF1 < QF2$.

\begin{figure}[!t]
\centering
\includegraphics[width=0.9\textwidth]{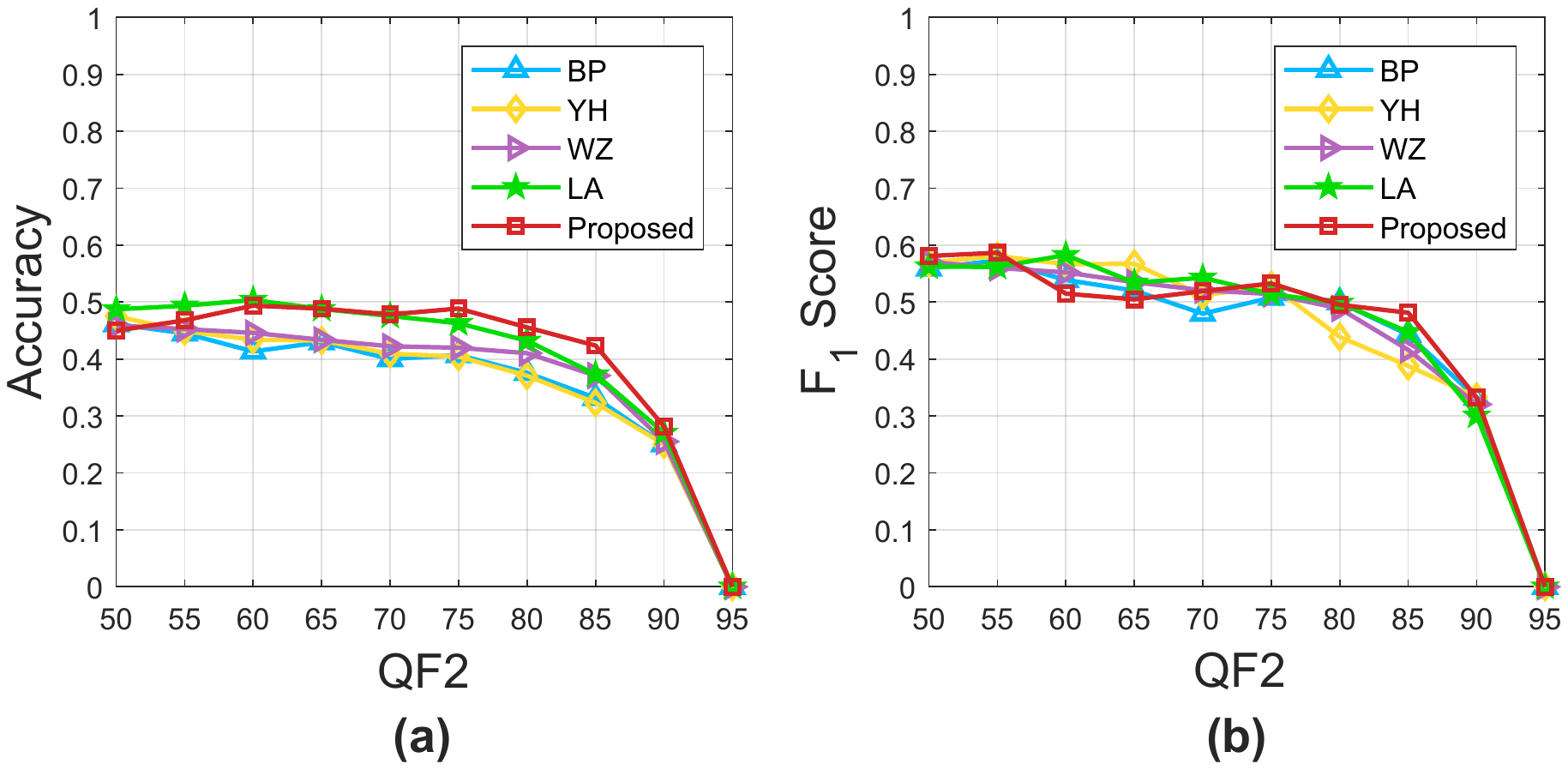}
\caption{\textcolor[rgb]{0,0,0}{Results achieved on the Synthetic dataset by the proposed method and the comparison methods of BP, YH, WZ, and LA when $QF1 \textgreater QF2$. (a)-(b) average values of the detection accuracy and $F_{1}Score$ corresponding to $QF2$.}}
\label{fig:accf1s_UCID_QF1greaterQF2}
\end{figure}

\begin{figure}[!t]
\centering
\includegraphics[width=0.9\textwidth]{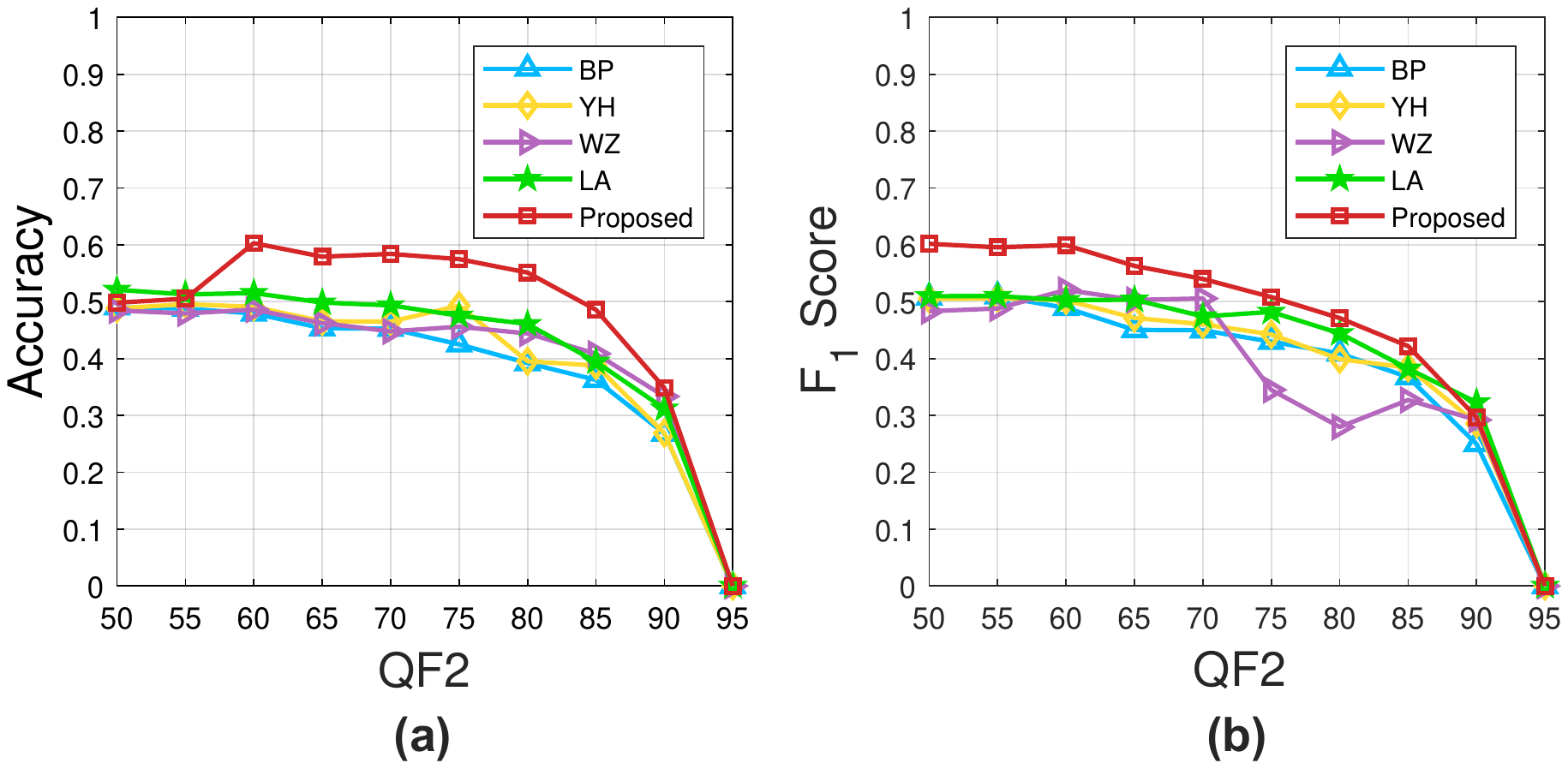}
\caption{\textcolor[rgb]{0,0,0}{Results achieved on the Florence dataset by the proposed method and the comparison methods of BP, YH, WZ, and LA when $QF1 > QF2$. (a)-(b) average values of the detection accuracy and $F_{1}Score$ corresponding to $QF2$.}}
\label{fig:accf1s_Dresden_QF1greaterQF2}
\end{figure}

In order to compare different approaches more intuitively, we also draw some figures. Because of the fact that detection performance is related to $QF1$ and $QF2$, we calculate the average values of $Acc$ and $F_{1}Score$ for all the images in different $QF1\in\{50,55,\dots,90,95\}$. Fig.~\ref{fig:accf1s_UCID} and Fig.~\ref{fig:accf1s_Dresden} illustrate the detection results on the Synthetic and Florence datasets, respectively.

It is evident that our method outperforms other methods in terms of accuracy and $F_{1}Score$ when $QF1 > QF2$ as well as $QF1 < QF2$. In general, the detection results of each method are significantly worse when $QF1 > QF2$ or $QF1=QF2$. Moreover, as we can see, the accuracy of each method is no more than $60\%$ in the case where $QF2 < 60$. As the compression quality factor ranges from $50$ to $95$, the $QF1$ of each image is almost higher than the $QF2$ in this case.

Fig.~\ref{fig:accf1s_UCID_QF1greaterQF2} and Fig.~\ref{fig:accf1s_Dresden_QF1greaterQF2} illustrate the average detection results when $QF1 > QF2$ in different $QF1\in\{50,55,\dots,90,95\}$. We find that our method has significantly higher accuracy and $F_{1}Score$ in this case. This greatly depends on the application of the discriminative module. Moreover, the multi-scale module also contributes to the overall improvement in detection.

These results show that our method has a notable superiority in double JPEG compression forensics. In addition, based on the slightly different trends in the results on the two datasets, our approach has more stable performance on the high-resolution dataset.

\begin{table}[!t] 
\renewcommand\arraystretch{1.5}
\centering
\tbl{\textcolor[rgb]{0,0,0}{Comparison of the proposed method and LA on Synthetic dataset.}\label{tab:LA_comparison}}{
\setlength{\tabcolsep}{1.85mm}
\begin{tabular}{|c|c|c|c|}
  \hline
  {Method}&{Average detection time}&{Feature dimension}&{Target area}\\
  \hline
  LA&570 ms/pic&94976&Global detection\\
  \hline
  Proposed&60 ms/pic&279&Accurate location\\
  \hline

\end{tabular}}
\end{table}

Table~\ref{tab:LA_comparison} shows the further comparison between our method and LA on Synthetic dataset. Since our method uses a much smaller number of features ($279$-D) than LA (nearly $100000$-D), the average detection time of our method is 60ms for each picture, almost 10 times faster than LA. Additionally, our method can accurately locate the tampered area of the JPEG image but LA is often used to determine the authenticity of the entire image.

\subsection{Qualitative Experiments}
\begin{figure}[!t]
\centering
\includegraphics[width=8 cm]{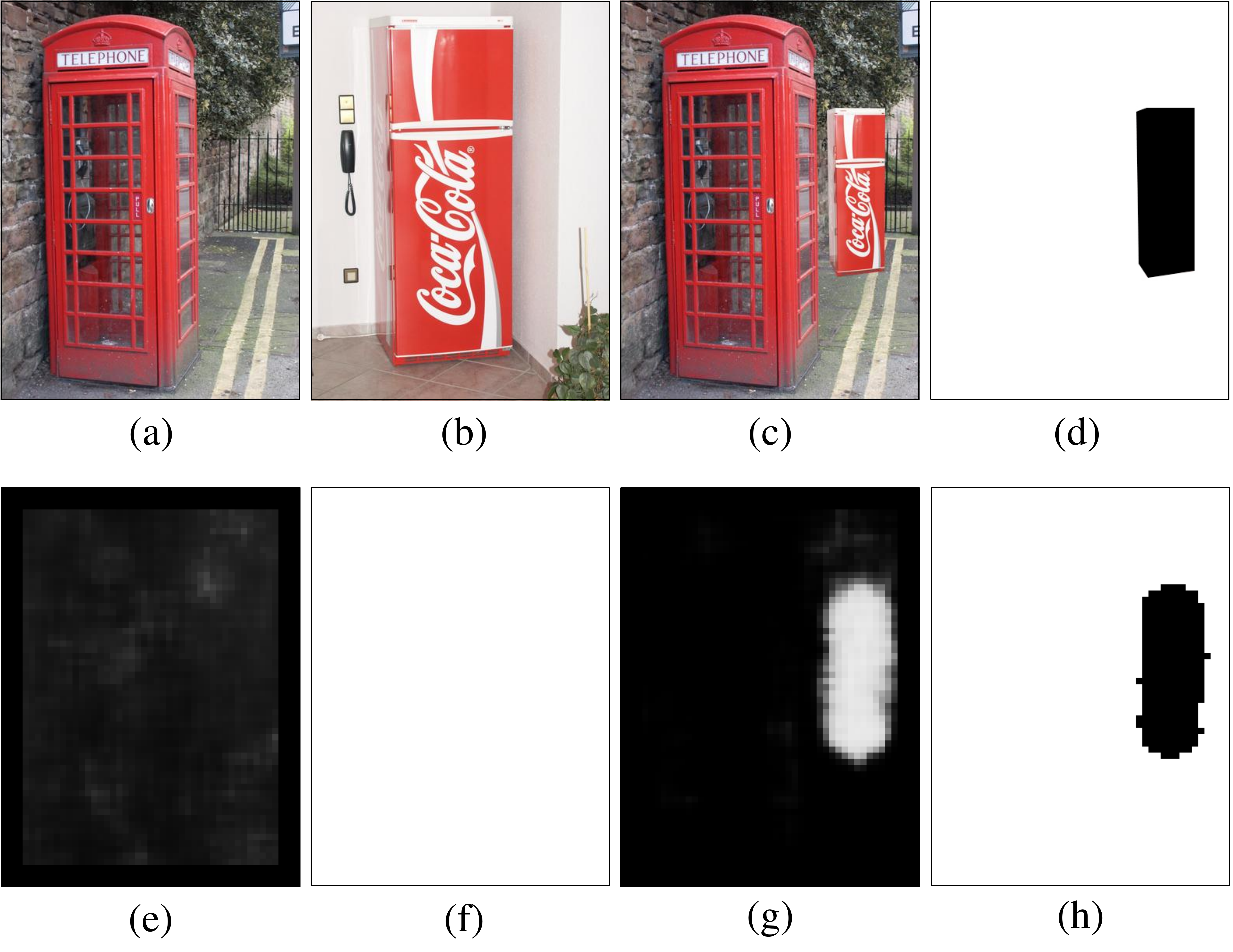}
\caption{A group of successful results of the proposed approach: (a) an original image 1, (b) an original image 2, (c) the tampered image 3 after narrowing, $QF1=60$, $QF2=90$, (d) the mask, (e) the classification probability map of image 1, (f) the detection result map of image 1, (g) the classification probability map of image 3, (h) the detection result map of image 3. In order to reflect the performance more intuitively, our results have not been subjected to any filtering.}
\label{fig:robust}
\end{figure}

To ensure the comprehensiveness of the experiment, we not only manually synthesize JPEG images using Adobe Photoshop, but also manipulate JPEG images automatically via Matlab. Fig.~\ref{fig:robust} shows the effectiveness and robustness of our method. We can observe from the last two columns that there are almost no tampered regions in image 1 that conform to our cognition. Although the refrigerator in the original image 2 is shrunken and inserted into another image, our method still yields a superior detection result.

\begin{figure}[!t]
\centering
\includegraphics[width=13.8 cm]{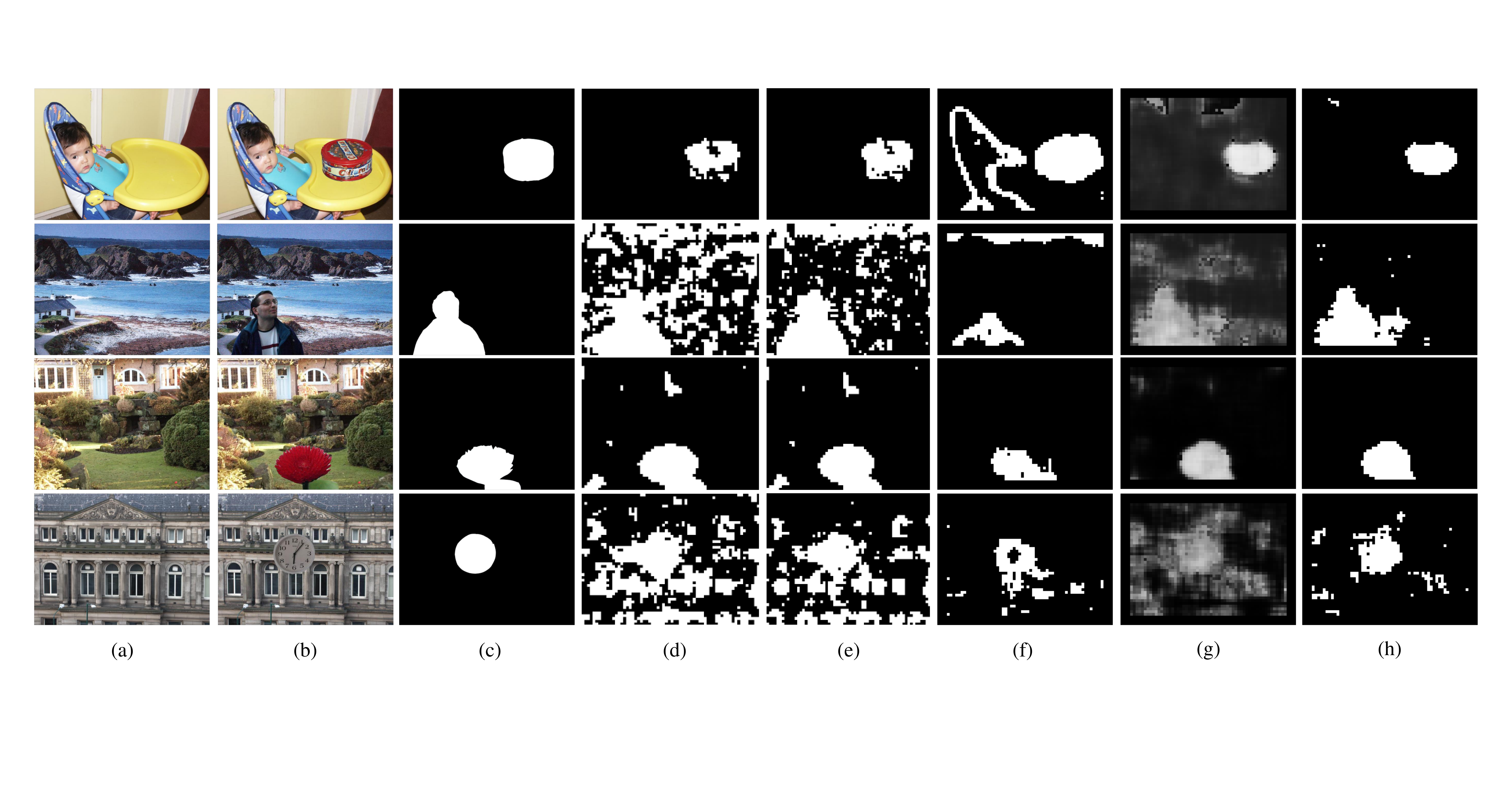}
\caption{Detection results on the Synthetic dataset: (a) original images, (b) tampered images compressed with $QF1=60$, $QF2=70$, and $QF1=80$, $QF2=70$ (two at the top which are human unrecognizable due to the reasonable semantic information they have), $QF1=50$, $QF2=100$, and $QF1=80$, $QF2=60$ (two at the bottom which are easy to recognize due to the abnormal semantic information then have), (c) tampering masks, (d)-(f) detection results of BP, YH, and WZ, (g) probability maps of our method, (h) detection results of our method.}
\label{fig:CompareUCID}
\end{figure}

\begin{figure}[!t]
\centering
\includegraphics[width=13.8 cm]{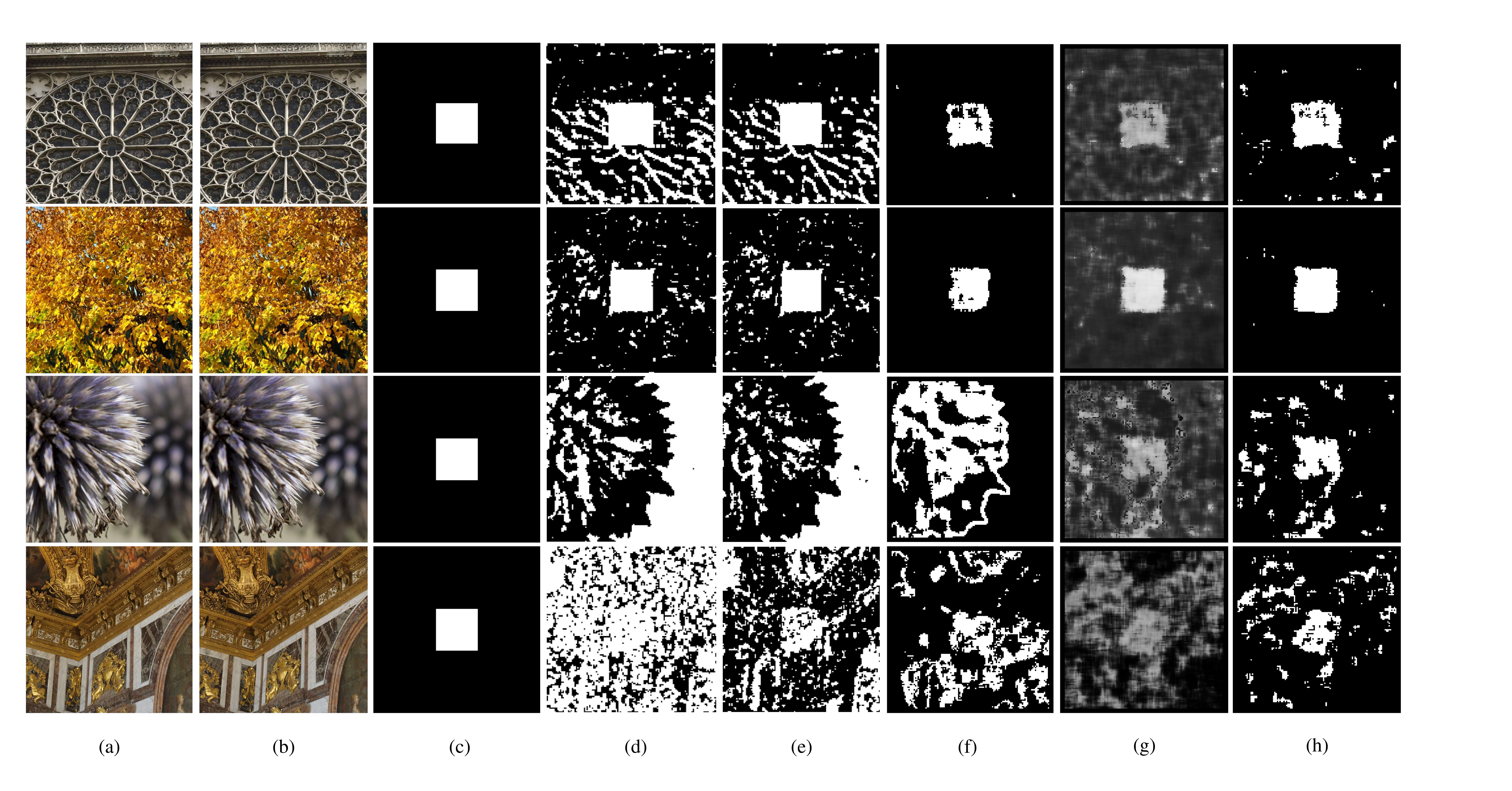}
\caption{Detection results on the Florence dataset: (a) original images, (b) tampered images compressed with $QF1=60$, $QF2=70$, and $QF1=50$, $QF2=100$ (two at the top), $QF1=80$, $QF2=70$, and $QF1=80$, $QF2=60$ (two at the bottom) are automatically synthesized by Matlab, the central square area of each image is replaced by a block with the same content, (c) tampering masks, (d)-(f) detection results of BP, YH, WZ, (g) probability maps of our method, (h) detection results of our method.}
\label{fig:CompareFlorence}
\end{figure}

Additional results compared with the methods of BP, YH, and WZ on different datasets are shown in Fig.~\ref{fig:CompareUCID} and Fig.~\ref{fig:CompareFlorence}. In order to ensure the diversity of the experiment, images from different datasets are selected and tampered in different ways.

\begin{table}[!t] 
\renewcommand\arraystretch{1.3}
\centering
\tbl{The accuracy of detection using different parameters.\label{tab:table3}}{ 
\setlength{\tabcolsep}{3.5mm}{
\begin{tabular}{c c c c c}
  \hline
  Dataset/Number of connections & 1 conv & \textbf{2 conv} & 3 conv & 4 conv \\
  \hline
  Synthetic & 0.687 & 0.690 & 0.682 & 0.685 \\
  Florence & 0.726 & 0.738 & 0.733 & 0.734 \\
  \hline
  \\
  \hline
  Dataset/Kernel size & $\mathbf{3 \times 1}$ & $5 \times 1$ & $7 \times 1$ & $9 \times 1$ \\
  \hline
  Synthetic & 0.690 & 0.682 & 0.681 & 0.679 \\
  Florence & 0.738 & 0.731 & 0.727 & 0.726 \\
  \hline
  \\
  \hline
  Dataset/Number of kernels & 50 & \textbf{100} & 150 & 200 \\
  \hline
  Synthetic & 0.677 & 0.690 & 0.688 & 0.689 \\
  Florence & 0.725 & 0.738 & 0.733 & 0.736 \\
  \hline
  \\
  \hline
  Dataset/Feature dimensions & 11 & 21 & \textbf{31} & 41 \\
  \hline
  Synthetic & 0.661 & 0.683 & 0.690 & 0.686 \\
  Florence & 0.682 & 0.704 & 0.738 & 0.740 \\
  \hline
  \\
  \hline
  Dataset/Network model & Model1 & Model2 & Model3 & \textbf{Model4} \\
  \hline
  Synthetic & 0.677 & 0.685 & 0.679 & 0.690 \\
  Florence & 0.723 & 0.734 & 0.728 & 0.738 \\
  \hline
\end{tabular}}}
\end{table}

Fig.~\ref{fig:CompareUCID} shows the detection results of artificially tampered images on the Synthetic dataset. It is evident that our method has fewer misclassified points. Simultaneously, our method is rarely influenced by interference information, such as the content of the sky in the second lines.

Fig.~\ref{fig:CompareFlorence} shows the detection results of automatically synthesized images on the Florence dataset. The results show that our method is rarely affected by the content of the image and performs better in cases of both $QF1 < QF2$ and $QF1 > QF2$, although all of the methods perform worse when $QF1 > QF2$. When $QF1 > QF2$, the superior performance of our method is derived from the discriminative module. The multi-scale features extracted by our networks can help to improve the classification effect and reduce the interference of the invalid information.

\subsection{Parameter Selection}
In this section, many experiments are implemented in order to reveal the relationship between global accuracy and parameter selection. Different DNN model parameters and structures are tested to construct better networks and specify the network parameters. Table~\ref{tab:table3} presents a comparison of the size of kernels, the number of convolutional layers, the number of kernels, the dimension of features, and the model composed of multiple networks which are trained by different scale data. Model1 to Model4 represent different network structures: 1) a network trained by $64 \times 64$ blocks; 2) two fused networks trained by $64 \times 64$ and $128 \times 128$ blocks; 3) two fused networks trained by $64 \times 64$ and $256 \times 256$ blocks; 4) three fused networks trained by $64 \times 64$, $128 \times 128$, and $256 \times 256$ blocks. The values in Table~\ref{tab:table3} which are in bold represent the parameters or structure we finally select.

\section{Conclusion}
This paper proposes a novel double JPEG compression forensics method based on deep multi-scale discriminative networks. The multi-scale features extracted by the multi-scale module derive more effective information from DCT coefficient histograms and achieve better performance in the tampering detection process. With reference to the statistical characteristics of the DQ effect, a discriminative module is also designed to capture the small difference between authentic and tampered images in tougher cases when $QF1 \textgreater QF2$. Finally, the automatic localization of specific tampered regions is realized. Extensive experimental results confirm that our MSD-Nets outperform several state-of-the-art methods on two public datasets.

In the future, it will be necessary for us to design a pretreatment process for filtering quantization noise. In addition, further efforts will be made to consider adding both the image content information and semantic information to assist in double JPEG compression forensics.


\bibliographystyle{ACM-Reference-Format-Journals}
\bibliography{refe}

\received{February 2007}{March 2009}{June 2009}


\end{document}